\documentclass[twoside]{article}
\usepackage[preprint]{aistats2027}

\usepackage[round]{natbib}
\renewcommand{\bibname}{References}
\renewcommand{\bibsection}{\subsubsection*{\bibname}}

\usepackage[utf8]{inputenc}
\usepackage[T1]{fontenc}
\usepackage{hyperref}
\usepackage{url}
\usepackage{booktabs}
\usepackage{multirow}
\usepackage{amsmath}
\usepackage{amsfonts}
\usepackage{amssymb}
\usepackage{microtype}
\usepackage{xcolor}
\usepackage{graphicx}

\begin{document}

\twocolumn[
\aistatstitle{NEST: Nested Embedding Representation for Multiscale Clustering}

\aistatsauthor{
Zhanqi Zhang$^{1}$
\And
Mikio Aoi$^{2,4,\dagger}$
\And
Gal Mishne$^{1,2,3,\dagger}$
}

\aistatsaddress{
$^{1}$Computer Science and Engineering, UC San Diego \\
$^{2}$Halıcıoğlu Data Science Institute, UC San Diego \\
$^{3}$Electrical and Computer Engineering, UC San Diego \\
$^{4}$Neurobiology, UC San Diego \\
$^{\dagger}$Equal contribution
}
]

\begin{abstract}
Real-world data often exhibits cluster structure at multiple resolutions, yet
existing methods typically learn a single partition or construct an explicit
hierarchy. Inspired by Matryoshka Representation Learning, we introduce
\textbf{N}ested \textbf{E}mbedding Repre\textbf{S}entation for
Mul\textbf{T}iscale Clustering (\textbf{NEST}), an unsupervised framework that
encodes coarse-to-fine structure within a single embedding. NEST jointly
optimizes clustering and separation across nested prefix subspaces, enabling
progressively finer distinctions as embedding dimensionality increases, with
rank and within-cluster regularization that keep the embedding from collapsing. Empirically, distances in the learned embedding follow the hierarchy
distance more closely than those of the input,
although the hierarchy is never observed during training. On synthetic
hierarchies, NEST recovers multiscale structure given only the number of
clusters per level. Experiments on vision, text and
single-cell data further demonstrate multiscale organization across real-world
representations. By making clustering granularity an intrinsic property of the
representation, NEST provides a unified approach to multiscale structure
discovery.

\end{abstract}

\section{Introduction}

\begin{figure*}[t]
\centering
\includegraphics[width=\textwidth]{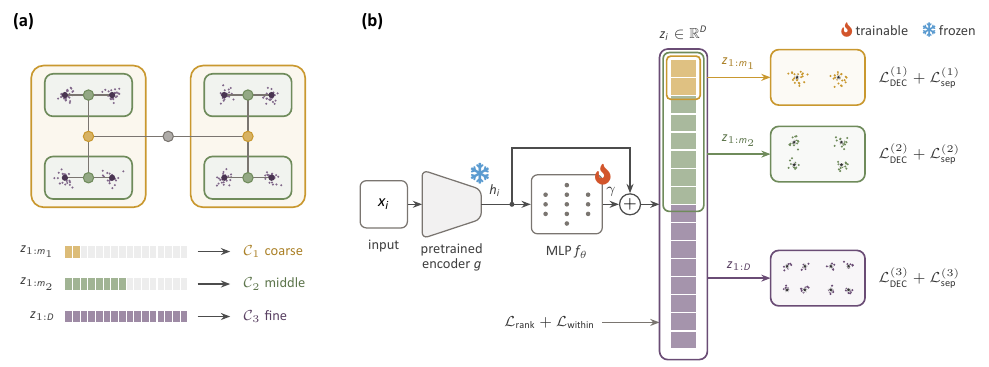}
\caption{\textbf{NEST.}
(a) Fine clusters $\mathcal{C}_3$ nest inside $\mathcal{C}_2$ and
$\mathcal{C}_1$; the {prefix}  $z_{1:m_\ell}$ resolves level $\ell$, with longer prefixes supporting progressively finer clustering.
(b) A trainable map $f_\theta$ adapts frozen features
$h_i=g(x_i)$ through a residual correction. Each {prefix} is optimized with
a level-specific clustering and separation objective
($\mathcal{L}^{(\ell)}_{\mathrm{DEC}}
+\mathcal{L}^{(\ell)}_{\mathrm{sep}}$), while the rank and
within-cluster regularizers $\mathcal{L}_{\mathrm{rank}}$ and
$\mathcal{L}_{\mathrm{within}}$ keep the embedding from collapsing.}
\label{fig:method}
\end{figure*}

Clustering is a fundamental tool for unsupervised structure discovery, yet
real-world data often admits meaningful organization at multiple levels of
resolution. Images, for example, may separate into broad semantic groups
such as animals and vehicles, and further into individual object categories.
Likewise, biological populations can be organized into broad cell classes,
subclasses, and increasingly specialized cell types. A useful representation
should therefore support structure discovery across several resolutions,
rather than only a single partition.

Classical clustering methods typically operate directly on a fixed feature space. Flat methods such as $k$-means \citep{kmeans} recover a partition at a specified resolution, while hierarchical methods recover nested partitions across multiple resolutions, either agglomeratively or divisively \citep{ward, bisectingkmeans}. Other approaches, such as spectral or hyperbolic embedding methods \citep{ng-spectral, Poincare}, first transform the data into a representation in which cluster structure is then identified. Across these settings, clustering quality depends on how well the representation preserves the distinctions required at each resolution. Clustering with deep learning \citep{dec, jule, DeepCluster} addresses the representation problem by jointly
learning embeddings and cluster assignments, but is typically optimized for
a single clustering resolution. Thus, hierarchical clustering provides
multiscale structure without representation learning, while deep clustering
provides representation learning without explicitly organizing multiple
resolutions.

We ask whether these two capabilities can be learned jointly, such that a single representation captures cluster structure at both coarse and fine resolutions simultaneously. Inspired by Matryoshka Representation Learning \citep{matryoshka}, we introduce
\textbf{N}ested \textbf{E}mbedding Repre\textbf{S}entation for
Mul\textbf{T}iscale Clustering (\textbf{NEST}). NEST learns a single embedding
whose nested prefixes are jointly optimized for progressively finer cluster
structure. Unlike conventional MRL, where prefixes of different lengths are
trained for the same downstream objective, for example, the same classification or retrieval task at different embedding sizes, NEST assigns each prefix a
different clustering resolution. Shorter prefixes capture coarser structure,
while longer prefixes capture progressively finer structure.

By encoding clustering resolution directly in the representation, NEST can learn multiscale cluster structure without access to a ground-truth hierarchy or explicit construction of a cluster tree during training. Instead, coarse-to-fine organization is learned through the
nested subspaces themselves. We show empirically that distances in
the learned embedding correlate more strongly with the hierarchy distance than
those of the input, connecting the learned representation to
hierarchical geometry. We validate NEST
on synthetic, visual, text, and single-cell data, where it recovers
coarse-to-fine cluster structure given only the number of clusters at each
resolution. Together, these results introduce a simple representation-based
approach to multiscale clustering and demonstrate that hierarchical
organization can emerge from nested subspaces without explicitly learning a
tree.

\section{Related Work}

\textbf{Hierarchical Clustering} organizes data into nested tree structures through bottom-up merging or top-down splitting \citep{ward,bisectingkmeans}.
Beyond such procedural constructions, \citet{dasgupta2015} formulated hierarchical clustering as optimization of a global tree objective over pairwise similarities.
Hyperbolic representation learning provides a complementary route for modeling tree-like structure, exploiting the geometry of hyperbolic space to embed hierarchical relations \citep{Poincare}. \citet{HypHC} connect these ideas by using hyperbolic embeddings to obtain a differentiable relaxation of Dasgupta's objective and recover a hierarchy from the learned geometry. More recent deep-learning-based hierarchical clustering methods couple representation learning with explicit tree construction. For example, \citet{deepect} jointly learn an autoencoder representation and a binary tree that expands through recursive splits, while \citet{CoHiClust} use contrastive learning to recover hierarchical cluster structure. In contrast to methods that represent hierarchy through explicit tree structures or hierarchical objectives, NEST encodes multiscale cluster structure directly in the representation, with nested embedding {prefixes} specialized to different resolutions.

\textbf{Representation Learning for Clustering.}
Learning feature representations has significantly improved clustering performance by producing embeddings that better separate semantic groups. Classical deep clustering approaches include \citep{dec, jule, DeepCluster}, which learn cluster structures jointly with representation learning. \citet{pica} maximize global partition confidence using a Partition Uncertainty Index to jointly optimize cluster separability and compactness. \citet{DCCM} mine multi-level correlations between samples using pseudo-graphs and pseudo-labels to guide representation learning. \citet{cc} jointly perform instance-level and cluster-level contrastive learning by treating feature rows as soft assignments and columns as cluster prototypes. Unlike these methods, which aim to learn a single clustering partition, our approach focuses on representations that support multiple clustering resolutions.

\textbf{Matryoshka Representation Learning } was introduced by \citet{matryoshka}, who
trained nested prefixes of an embedding to remain useful at different
dimensionalities. \citet{N2F2} learn a nested representation for 3D scene
understanding, but their coarse-to-fine structure is explicitly provided during
training.
\citet{mrl-nrp} train multilingual Matryoshka embeddings using labeled pairs of
news articles with graded semantic similarity, applying a contrastive loss at
different prefix lengths so that the prefixes encode progressively finer
notions of news similarity. They then apply a separate hierarchical clustering
algorithm to the learned embeddings. FRANCA \citep{Franca} is close to our setting in that it also uses nested prefixes with progressively finer prototype assignments. However, FRANCA uses these assignments to learn visual representations, whereas NEST treats the clusterings themselves as the target output, directly optimizing multiscale clustering on a fixed input representation.

\section{Nested Clustering}
\label{sec:method}

We formulate multiscale clustering as learning a single nested representation that supports clustering at multiple resolutions. NEST uses a Matryoshka-style embedding in which progressively longer prefixes encode increasingly fine structure (Figure~\ref{fig:method}a).

\textbf{Problem setup.}
Let $\mathcal{D}=\{x_i\}_{i=1}^{N}$ be an unlabeled dataset, and $L$ denote the number of clustering resolutions. For each level $\ell\in\{1,\ldots,L\}$, we specify a target number of clusters $K_\ell$, ordered from coarse to fine,
\[
K_1 < K_2 < \cdots < K_L.
\]
We denote the clustering induced at level $\ell$ by
\[
\mathcal{C}_\ell
=
\{C_{\ell,1},\ldots,C_{\ell,K_\ell}\}.
\]

Given these resolutions, the only multiscale information supplied to NEST is the target cluster count \(K_\ell\) at each resolution; no parent-child assignments, tree structure, or ground-truth hierarchy are used during training.

\textbf{Nested representation.}
Each sample is first mapped by a frozen feature extractor $g$ (e.g., a pretrained encoder, or a fixed projection such as PCA) to features
$h_i=g(x_i)\in\mathbb{R}^{D}$. NEST learns a residual map on these features,
\begin{equation}
z_i=O h_i+\gamma\, f_\theta(h_i)\in\mathbb{R}^{D},
\label{eq:residual}
\end{equation}
where $f_\theta$ is a trainable MLP (Figure~\ref{fig:method}b) and $\gamma>0$ sets the scale of the learned correction. $O$ is a fixed orthogonal $D\times D$ matrix, so NEST keeps the encoder's width. The MLP's last layer is initialized to zero so at initialization, $z_i=O h_i$ has identical pairwise distances of the input. On the real datasets $O$ rotates $h_i$ into its principal axes, computed once from the features, ordering the coordinates by variance so that each prefix starts from the leading directions. On the synthetic data $O=I$ (Appendix~\ref{app:hyper}).

We choose nested prefix dimensions $m_1 < \cdots < m_L=D $, and define the representation used at level $\ell$ as
\[
z_i^{(\ell)}
:=
z_{i,:m_\ell}
\in\mathbb{R}^{m_\ell}.
\]
Thus, coarse clustering operates on a shorter prefix, while finer levels retain the coarse coordinates and progressively add further coordinates.

\textbf{Level-wise clustering.}
For each level $\ell$, NEST learns $K_\ell$ trainable cluster centers
\[
\left\{
\mu_k^{(\ell)}
\right\}_{k=1}^{K_\ell},
\qquad
\mu_k^{(\ell)}\in\mathbb{R}^{m_\ell}.
\]

Each center represents one cluster at that resolution. A point is assigned to a cluster based on its distance to these centers in the corresponding prefix space. At each resolution, NEST applies the Deep Embedded Clustering (DEC) objective \citep{dec} to the corresponding prefix representation.

For sample $i$ and center $k$ at level $\ell$, we compute the Euclidean distance between the level-specific prefix $z_i^{(\ell)}$ and cluster center $\mu_k^{(\ell)}$, scaled per level by $a_\ell/\widehat s_\ell$, where $a_\ell>0$ is the level weight and $\widehat s_\ell$ the initial center spacing (both defined below):
\begin{equation}
\delta_{ik}^{(\ell)}
=
\frac{a_\ell}{\widehat s_\ell}
\left\|
z_i^{(\ell)}
-
\mu_k^{(\ell)}
\right\|_2 ,
\label{eq:calibrated-distance}
\end{equation}

These rescaled distances measure how far sample $i$ is from cluster
$k$ at resolution $\ell$.

Following DEC \citep{dec}, we convert these distances into Student-$t$ soft assignments over the $K_\ell$ clusters, by assigning a nonnegative weight $q_{ik}^{(\ell)}$ to cluster $k$, with larger weights for centers closer to $z_i^{(\ell)}$
\begin{equation}
q_{ik}^{(\ell)}
=
\frac{
\left(1+\left(\delta_{ik}^{(\ell)}\right)^2\right)^{-1}
}{
\sum_{j=1}^{K_\ell}
\left(1+\left(\delta_{ij}^{(\ell)}\right)^2\right)^{-1}
}.
\label{eq:soft-assignment}
\end{equation}

To give more weight to confident assignments while preventing large clusters from dominating the target distribution, the soft assignments are sharpened and reweighted by cluster frequency. Let
\[
f_k^{(\ell)}
=
\sum_i q_{ik}^{(\ell)},
\]
where $f_k^{(\ell)}$ is the total soft-assignment mass of cluster $k$. The sharpened target distribution is
\begin{equation}
p_{ik}^{(\ell)}
=
\frac{
\left(q_{ik}^{(\ell)}\right)^2/f_k^{(\ell)}
}{
\sum_{j=1}^{K_\ell}
\left(q_{ij}^{(\ell)}\right)^2/f_j^{(\ell)}
},
\label{eq:dec-target}
\end{equation}
where $P^{(\ell)}$ is computed from the current assignments but treated as a fixed target during optimization. Gradients are not propagated through $P^{(\ell)}$.

The clustering objective at level $\ell$ is then the KL divergence between the sharpened target distribution $P^{(\ell)}$ and the current soft assignments
$Q^{(\ell)}$:
\begin{equation}
\mathcal{L}_{\mathrm{DEC}}^{(\ell)}
=
\mathrm{KL}
\left(
P^{(\ell)}
\,\Vert\,
Q^{(\ell)}
\right).
\label{eq:dec-loss}
\end{equation}

\textbf{Center-separation penalty.}
The DEC objective encourages samples to concentrate around cluster centers but does not explicitly keep different cluster centers separated. As a result, multiple centers can move toward the same location, causing different clusters to collapse onto one another.

We therefore add a separation penalty that keeps different cluster centers at least a margin $\rho_\ell$ apart. Specifically, at level $\ell$, let $k$ and $k'$ index two different cluster centers among the $K_\ell$ centers. We define
\[
\mathcal{E}_\ell
=
\{(k,k'):1\leq k<k'\leq K_\ell\}
\]
as the set of all unordered pairs of different cluster centers at that level. The separation loss is
\begin{equation}
\mathcal{L}_{\mathrm{sep}}^{(\ell)}
=
\frac{1}{|\mathcal{E}_\ell|}
\sum_{(k,k')\in\mathcal{E}_\ell}
\left[
\rho_\ell
-
\left\|
\mu_k^{(\ell)}
-
\mu_{k'}^{(\ell)}
\right\|_2
\right]_+ ,
\label{eq:loss-sep}
\end{equation}
where $\rho_\ell$ is the required minimum separation between centers at level $\ell$, and $[u]_+=\max(u,0)$ denotes the positive-part function which returns $u$ when $u>0$ and $0$ otherwise.

Thus, a pair of centers is penalized only when their distance is smaller than $\rho_\ell$, and centers already separated by at least $\rho_\ell$ receive no penalty.

\textbf{Setting scale parameters.}
\label{sec:scale-calibration}
We now define the level-specific distance scale $\hat{s}_\ell$ and separation margin $\rho_\ell$ introduced above. The cluster centers are initialized independently at each level by $k$-means (\texttt{n\_init}$=10$) on the initial corresponding prefix $z^{(\ell)}$. For the full resolution $\ell=L$, this is equivalent to $k$-means in the input feature space $h_i$. From this initialization, we compute two level-specific distance statistics.

First, $\widehat{s}_\ell$ is the minimum Euclidean distance between initialized cluster centers at level $\ell$, measuring the separation of the closest pair of clusters.
Thus the distances used in the DEC soft assignments are normalized relative to the initial spacing between cluster centers.

Second, $\widehat{r}_\ell$ is the mean Euclidean distance from each sample to the center of the cluster assigned to it by the initial $k$-means solution. It measures the typical initial within-cluster radius at level $\ell$. We use it to set the separation margin
\[
\rho_\ell
=
4\widehat{r}_\ell,
\]
so that two learned centers are penalized when they become closer than four times the initial within-cluster radius. The factor $4$ is a heuristic taken from a worst-case argument (Appendix~\ref{app:margin-geometry}).

\begin{table*}[t]
\centering
\footnotesize
\setlength{\tabcolsep}{1pt}
\renewcommand{\arraystretch}{1.12}

\caption{\textbf{Clustering on synthetic data.}
Mean$\pm$s.d. over 5 seeds, with each seed drawing a new dataset. Best in bold and second best underlined.}
\label{tab:synth}

\begin{tabular*}{\textwidth}{
@{\extracolsep{\fill}}
l c
cccc
cccc
cccc
@{}}
\toprule
& &
\multicolumn{4}{c}{NMI} &
\multicolumn{4}{c}{ARI} &
\multicolumn{4}{c}{ACC} \\
\cmidrule(lr){3-6}
\cmidrule(lr){7-10}
\cmidrule(lr){11-14}

Dataset & $K$
& $k$-means & Ward & HDB & NEST
& $k$-means & Ward & HDB & NEST
& $k$-means & Ward & HDB & NEST \\
\midrule

\multirow{3}{*}{\shortstack[l]{Nested\\GMM}}
& $4$
& $\mathbf{1.00}{\scriptstyle\pm.00}$ & $\mathbf{1.00}{\scriptstyle\pm.00}$ & $\mathbf{1.00}{\scriptstyle\pm.00}$ & $\mathbf{1.00}{\scriptstyle\pm.00}$
& $\mathbf{1.00}{\scriptstyle\pm.00}$ & $\mathbf{1.00}{\scriptstyle\pm.00}$ & $\mathbf{1.00}{\scriptstyle\pm.00}$ & $\mathbf{1.00}{\scriptstyle\pm.00}$
& $\mathbf{1.00}{\scriptstyle\pm.00}$ & $\mathbf{1.00}{\scriptstyle\pm.00}$ & $\mathbf{1.00}{\scriptstyle\pm.00}$ & $\mathbf{1.00}{\scriptstyle\pm.00}$ \\

& $16$
& $\underline{.99}{\scriptstyle\pm.01}$ & $\underline{.99}{\scriptstyle\pm.00}$ & $.67{\scriptstyle\pm.00}$ & $\mathbf{1.00}{\scriptstyle\pm.00}$
& $.95{\scriptstyle\pm.04}$ & $\underline{.99}{\scriptstyle\pm.00}$ & $.33{\scriptstyle\pm.00}$ & $\mathbf{1.00}{\scriptstyle\pm.00}$
& $.94{\scriptstyle\pm.05}$ & $\underline{.99}{\scriptstyle\pm.00}$ & $.25{\scriptstyle\pm.00}$ & $\mathbf{1.00}{\scriptstyle\pm.00}$ \\

& $64$
& $\underline{.67}{\scriptstyle\pm.00}$ & $\underline{.67}{\scriptstyle\pm.00}$ & $.50{\scriptstyle\pm.00}$ & $\mathbf{.82}{\scriptstyle\pm.01}$
& $\underline{.25}{\scriptstyle\pm.00}$ & $\underline{.25}{\scriptstyle\pm.00}$ & $.09{\scriptstyle\pm.00}$ & $\mathbf{.41}{\scriptstyle\pm.02}$
& $\underline{.28}{\scriptstyle\pm.00}$ & $\underline{.28}{\scriptstyle\pm.00}$ & $.06{\scriptstyle\pm.00}$ & $\mathbf{.29}{\scriptstyle\pm.04}$ \\

\midrule

\multirow{3}{*}{\shortstack[l]{Balanced\\tree}}
& $3$
& $\underline{.75}{\scriptstyle\pm.01}$ & $.73{\scriptstyle\pm.01}$ & $.00{\scriptstyle\pm.00}$ & $\mathbf{.76}{\scriptstyle\pm.01}$
& $\underline{.71}{\scriptstyle\pm.01}$ & $.70{\scriptstyle\pm.01}$ & $.00{\scriptstyle\pm.00}$ & $\mathbf{.72}{\scriptstyle\pm.02}$
& $\underline{.74}{\scriptstyle\pm.01}$ & $.73{\scriptstyle\pm.01}$ & $.48{\scriptstyle\pm.00}$ & $\mathbf{.75}{\scriptstyle\pm.02}$ \\

& $7$
& $\underline{.79}{\scriptstyle\pm.07}$ & $.67{\scriptstyle\pm.06}$ & $.05{\scriptstyle\pm.07}$ & $\mathbf{.95}{\scriptstyle\pm.08}$
& $\underline{.72}{\scriptstyle\pm.10}$ & $.64{\scriptstyle\pm.06}$ & $.00{\scriptstyle\pm.01}$ & $\mathbf{.94}{\scriptstyle\pm.12}$
& $\underline{.71}{\scriptstyle\pm.07}$ & $.70{\scriptstyle\pm.04}$ & $.25{\scriptstyle\pm.03}$ & $\mathbf{.94}{\scriptstyle\pm.11}$ \\

& $15$
& $\underline{.52}{\scriptstyle\pm.05}$ & $.46{\scriptstyle\pm.05}$ & $.05{\scriptstyle\pm.05}$ & $\mathbf{.77}{\scriptstyle\pm.02}$
& $\underline{.26}{\scriptstyle\pm.03}$ & $.23{\scriptstyle\pm.04}$ & $.00{\scriptstyle\pm.00}$ & $\mathbf{.50}{\scriptstyle\pm.02}$
& $\underline{.34}{\scriptstyle\pm.03}$ & $.33{\scriptstyle\pm.03}$ & $.11{\scriptstyle\pm.01}$ & $\mathbf{.48}{\scriptstyle\pm.03}$ \\

\midrule

\multirow{3}{*}{\shortstack[l]{Random\\branching}}
& $4$
& $\underline{.75}{\scriptstyle\pm.01}$ & $\mathbf{.76}{\scriptstyle\pm.01}$ & $.00{\scriptstyle\pm.00}$ & $.60{\scriptstyle\pm.13}$
& $\underline{.52}{\scriptstyle\pm.02}$ & $\mathbf{.57}{\scriptstyle\pm.04}$ & $.00{\scriptstyle\pm.00}$ & $.39{\scriptstyle\pm.13}$
& $.69{\scriptstyle\pm.05}$ & $\mathbf{.76}{\scriptstyle\pm.06}$ & $\underline{.72}{\scriptstyle\pm.00}$ & $.63{\scriptstyle\pm.08}$ \\

& $13$
& $\underline{.75}{\scriptstyle\pm.04}$ & $.53{\scriptstyle\pm.02}$ & $.05{\scriptstyle\pm.07}$ & $\mathbf{.97}{\scriptstyle\pm.02}$
& $\underline{.62}{\scriptstyle\pm.06}$ & $.43{\scriptstyle\pm.04}$ & $.01{\scriptstyle\pm.01}$ & $\mathbf{.91}{\scriptstyle\pm.07}$
& $\underline{.61}{\scriptstyle\pm.05}$ & $.55{\scriptstyle\pm.05}$ & $.17{\scriptstyle\pm.03}$ & $\mathbf{.92}{\scriptstyle\pm.06}$ \\

& $52$
& $\underline{.45}{\scriptstyle\pm.02}$ & $.36{\scriptstyle\pm.01}$ & $.07{\scriptstyle\pm.03}$ & $\mathbf{.70}{\scriptstyle\pm.02}$
& $\underline{.13}{\scriptstyle\pm.02}$ & $\underline{.13}{\scriptstyle\pm.01}$ & $.00{\scriptstyle\pm.00}$ & $\mathbf{.27}{\scriptstyle\pm.04}$
& $\underline{.18}{\scriptstyle\pm.01}$ & $.16{\scriptstyle\pm.00}$ & $.05{\scriptstyle\pm.01}$ & $\mathbf{.25}{\scriptstyle\pm.02}$ \\

\bottomrule
\end{tabular*}
\end{table*}

\textbf{Degeneracy regularization.}
The clustering objective can lead to two degenerate solutions: collapse onto a
low-dimensional subspace, and contraction of clusters to near-point masses.
These are different failures: the first is a global loss of directions, which
$\mathcal{L}_{\mathrm{rank}}$ addresses, and the second is a local contraction
within clusters, which $\mathcal{L}_{\mathrm{within}}$ addresses.

The rank regularization discourages
the representation from collapsing onto a small number of directions
\begin{equation}
\mathcal{L}_{\mathrm{rank}}
=
\frac{1}{L}\sum_{\ell=1}^{L}
\left[1-\frac{e_\ell}{\tau_{\mathrm r}e_\ell^0}\right]_+^2 .
\label{eq:loss-rank}
\end{equation}
Here $e_\ell=\operatorname{erank}(B^{(\ell)})$ is the effective rank of
$B^{(\ell)}\in\mathbb{R}^{N\times(m_\ell-m_{\ell-1})}$, whose rows contain the coordinates introduced at level $\ell$, $z_{i,\,m_{\ell-1}+1:m_\ell}$, with $m_0=0$.
\begin{equation}
\operatorname{erank}(Z)=\exp\Bigl(-\sum_j p_j\log p_j\Bigr),
\label{eq:erank}
\end{equation}
where $p_j=\sigma_j^2/\sum_k\sigma_k^2$ and $\sigma_j$ are the singular values
of the column-centered matrix $Z$.
The reference $e_\ell^0$ is the same quantity for the initial embedding
$z=Oh$.

The cluster spread regularization keeps clusters from contracting to near-point masses, preserving non-degenerate within-cluster variation; when applied at a coarser level, that variation is also what finer partitions are built from. At level $\ell$,
\begin{equation}
\mathcal{L}_{\mathrm{within}}
=
\left[1-\frac{w^{(\ell)}}{\tau_{\mathrm w}w_0^{(\ell)}}\right]_+^2 ,
\label{eq:loss-within}
\end{equation}
and the loss is averaged over the levels at which it is applied (Table~\ref{tab:hyper}).
Here $w^{(\ell)}$ is the within-cluster spread at level $\ell$, the mean
squared distance of a point's prefix to the mean of its assigned
cluster,
\begin{equation}
w^{(\ell)}=\frac{1}{N}\sum_{i=1}^{N}
\bigl\|z_i^{(\ell)}-\bar z^{(\ell)}_{c_i}\bigr\|^2,
\label{eq:within-spread}
\end{equation}
where $c_i=\arg\min_k\delta_{ik}^{(\ell)}$ is the cluster assigned to point
$i$, $\bar z^{(\ell)}_k$ is the mean of $z_i^{(\ell)}$ over the points
assigned to cluster $k$, and $w_0^{(\ell)}$ is the same quantity at
initialization.

The activation fractions $\tau_{\mathrm r}$ and $\tau_{\mathrm w}$ determine
when each soft penalty begins to increase. Otherwise the terms remain zero,
allowing the embedding to move freely away from $h$.

\textbf{NEST objective.}
All levels are optimized jointly with
\begin{equation}
\sum_{\ell=1}^{L}
\bar a_\ell\,
\bigl(\mathcal{L}_{\mathrm{DEC}}^{(\ell)}+\mathcal{L}_{\mathrm{sep}}^{(\ell)}\bigr)
+
\lambda_{\mathrm{rank}}
\mathcal{L}_{\mathrm{rank}}
+
\lambda_{\mathrm{within}}
\mathcal{L}_{\mathrm{within}} .
\label{eq:nest-objective}
\end{equation}
Here $\bar a_\ell=a_\ell\big/\bigl(\tfrac1L\sum_{k}a_k\bigr)$ is the level weight of
\eqref{eq:calibrated-distance} normalized to mean $1$, so the same $a_\ell$ balances
the resolutions in the distance and in the objective, and $\lambda_{\mathrm{rank}}$
and $\lambda_{\mathrm{within}}$ weight the two regularizers.

\begin{table*}[t]
\centering
\footnotesize
\caption{\textbf{Clustering on real data.} Randomized methods and NEST report mean\textpm{}s.d.\ over the same $3$ seeds; Ward and Avg-L are deterministic. Best in bold, second best underlined.}
\label{tab:real-results}

\begingroup
\setlength{\tabcolsep}{3pt}
\renewcommand{\arraystretch}{0.88}

\begin{tabular*}{\textwidth}{@{\extracolsep{\fill}}l ccc ccc@{}}
\toprule  & \multicolumn{3}{c}{$K=20$} & \multicolumn{3}{c}{$K=100$} \\
\cmidrule(lr){2-4}
\cmidrule(lr){5-7}
\textbf{CIFAR-100} & NMI & ARI & ACC & NMI & ARI & ACC \\
\midrule
k-means & .33{\scriptsize\textpm.00} & .10{\scriptsize\textpm.01} & .30{\scriptsize\textpm.01} & .48{\scriptsize\textpm.00} & .10{\scriptsize\textpm.00} & \underline{.28}{\scriptsize\textpm.00} \\
Ward & .33 & .08 & .28 & .47 & .09 & .27 \\
Avg-L (Euc) & .02 & .00 & .06 & .13 & .00 & .06 \\
HypHC & .07{\scriptsize\textpm.02} & .02{\scriptsize\textpm.01} & .10{\scriptsize\textpm.01} & .23{\scriptsize\textpm.01} & .01{\scriptsize\textpm.00} & .06{\scriptsize\textpm.00} \\
Bisect.\ k-means & .27{\scriptsize\textpm.00} & .10{\scriptsize\textpm.00} & .24{\scriptsize\textpm.00} & .44{\scriptsize\textpm.00} & .09{\scriptsize\textpm.00} & .22{\scriptsize\textpm.00} \\
DeepECT & \underline{.38}{\scriptsize\textpm.02} & \underline{.21}{\scriptsize\textpm.02} & \underline{.36}{\scriptsize\textpm.02} & \underline{.49}{\scriptsize\textpm.01} & \underline{.14}{\scriptsize\textpm.01} & .27{\scriptsize\textpm.01} \\
\textbf{NEST} & \textbf{.43}{\scriptsize\textpm.01} & \textbf{.24}{\scriptsize\textpm.01} & \textbf{.42}{\scriptsize\textpm.01} & \textbf{.56}{\scriptsize\textpm.00} & \textbf{.22}{\scriptsize\textpm.00} & \textbf{.35}{\scriptsize\textpm.01} \\
\end{tabular*}
\vspace{3pt}

\begin{tabular*}{\textwidth}{@{\extracolsep{\fill}}l ccc ccc@{}}
 & \multicolumn{3}{c}{$K=6$} & \multicolumn{3}{c}{$K=20$} \\
\cmidrule(lr){2-4}
\cmidrule(lr){5-7}
\textbf{20 Newsgroups} & NMI & ARI & ACC & NMI & ARI & ACC \\
\midrule
k-means & \underline{.44}{\scriptsize\textpm.00} & .37{\scriptsize\textpm.00} & \underline{.63}{\scriptsize\textpm.00} & \underline{.53}{\scriptsize\textpm.00} & \underline{.37}{\scriptsize\textpm.01} & \underline{.54}{\scriptsize\textpm.01} \\
Ward & .43 & .37 & .61 & .50 & .32 & .52 \\
Avg-L (Euc) & .25 & .18 & .40 & .44 & .21 & .27 \\
HypHC & .23{\scriptsize\textpm.00} & .17{\scriptsize\textpm.00} & .38{\scriptsize\textpm.00} & .17{\scriptsize\textpm.00} & .06{\scriptsize\textpm.00} & .10{\scriptsize\textpm.00} \\
Bisect.\ k-means & .40{\scriptsize\textpm.01} & .33{\scriptsize\textpm.01} & .56{\scriptsize\textpm.02} & .46{\scriptsize\textpm.01} & .30{\scriptsize\textpm.00} & .47{\scriptsize\textpm.02} \\
DeepECT & \underline{.44}{\scriptsize\textpm.02} & \underline{.42}{\scriptsize\textpm.05} & .61{\scriptsize\textpm.05} & .47{\scriptsize\textpm.00} & .31{\scriptsize\textpm.01} & .43{\scriptsize\textpm.01} \\
\textbf{NEST} & \textbf{.47}{\scriptsize\textpm.02} & \textbf{.44}{\scriptsize\textpm.02} & \textbf{.68}{\scriptsize\textpm.01} & \textbf{.54}{\scriptsize\textpm.01} & \textbf{.41}{\scriptsize\textpm.01} & \textbf{.57}{\scriptsize\textpm.02} \\
\end{tabular*}
\vspace{3pt}

\begin{tabular*}{\textwidth}{@{\extracolsep{\fill}}l ccc ccc ccc@{}}
 & \multicolumn{3}{c}{$K=4$} & \multicolumn{3}{c}{$K=23$} & \multicolumn{3}{c}{$K=133$} \\
\cmidrule(lr){2-4}
\cmidrule(lr){5-7}
\cmidrule(lr){8-10}
\textbf{Tasic 2018} & NMI & ARI & ACC & NMI & ARI & ACC & NMI & ARI & ACC \\
\midrule
k-means & \underline{.83}{\scriptsize\textpm.00} & \underline{.79}{\scriptsize\textpm.00} & \underline{.77}{\scriptsize\textpm.00} & \textbf{.89}{\scriptsize\textpm.00} & \textbf{.74}{\scriptsize\textpm.01} & \underline{.77}{\scriptsize\textpm.01} & \underline{.84}{\scriptsize\textpm.00} & .50{\scriptsize\textpm.02} & .56{\scriptsize\textpm.01} \\
Ward & .82 & \underline{.79} & \underline{.77} & \textbf{.89} & \underline{.71} & .75 & \underline{.84} & .53 & \underline{.59} \\
Avg-L (Euc) & .41 & .18 & .55 & .83 & .56 & .69 & .83 & \underline{.55} & .57 \\
HypHC & \textbf{.85}{\scriptsize\textpm.00} & \textbf{.89}{\scriptsize\textpm.00} & \textbf{.94}{\scriptsize\textpm.00} & .41{\scriptsize\textpm.00} & .18{\scriptsize\textpm.00} & .26{\scriptsize\textpm.00} & .20{\scriptsize\textpm.00} & .03{\scriptsize\textpm.00} & .07{\scriptsize\textpm.00} \\
Bisect.\ k-means & .82{\scriptsize\textpm.00} & \underline{.79}{\scriptsize\textpm.00} & \underline{.77}{\scriptsize\textpm.00} & \underline{.87}{\scriptsize\textpm.00} & .70{\scriptsize\textpm.02} & .73{\scriptsize\textpm.02} & .81{\scriptsize\textpm.00} & .44{\scriptsize\textpm.00} & .50{\scriptsize\textpm.00} \\
DeepECT & .65{\scriptsize\textpm.02} & .53{\scriptsize\textpm.04} & .62{\scriptsize\textpm.04} & \underline{.87}{\scriptsize\textpm.01} & .67{\scriptsize\textpm.01} & .68{\scriptsize\textpm.02} & \underline{.84}{\scriptsize\textpm.00} & \underline{.55}{\scriptsize\textpm.04} & .58{\scriptsize\textpm.02} \\
\textbf{NEST} & \underline{.83}{\scriptsize\textpm.00} & \underline{.79}{\scriptsize\textpm.00} & \underline{.77}{\scriptsize\textpm.00} & \textbf{.89}{\scriptsize\textpm.00} & \textbf{.74}{\scriptsize\textpm.01} & \textbf{.78}{\scriptsize\textpm.01} & \textbf{.85}{\scriptsize\textpm.00} & \textbf{.64}{\scriptsize\textpm.01} & \textbf{.66}{\scriptsize\textpm.01} \\
\bottomrule
\end{tabular*}

\endgroup
\end{table*}

\section{Experiments}
\label{sec:experiments}

We evaluate NEST on three synthetic and three real datasets. Each dataset provides nested labels, with the number of clusters at level $\ell$ set to the number of classes at that level $K=(K_1,\ldots,K_L)$. NEST is trained without label supervision and compared against six baselines on the real datasets and three on the synthetic datasets, using clustering metrics at every level. Across datasets, we keep most architectural and optimization settings fixed. The complete settings are reported in
Table~\ref{tab:hyper}, with sensitivity to these choices evaluated in
Appendix~\ref{app:sensitivity}.

\subsection{Datasets}
\label{sec:datasets}

\textbf{Synthetic datasets}
were generated with known trees.
The \textbf{nested GMM} is a three-level Gaussian mixture, $K=(4,16,64)$,
in which each cluster of one level splits into four at the next.
The \textbf{balanced tree}, $K=(3,7,15)$, places cluster centers by a
top-down walk on a balanced binary tree, so that Euclidean distance between
centers tracks tree distance.
The \textbf{random-branching tree}, $K=(4,13,52)$, does the same on a tree
with a random number of children per node, giving coarse classes of very
unequal size.

For all three datasets, level-specific hierarchical signals are first placed in separate coordinate blocks and orthogonalized. The resulting signal vector is then rotated by a fixed random orthonormal matrix, after which independent nuisance dimensions are appended.
Generator parameters are in Appendix~\ref{app:data}.

\textbf{CIFAR-100.}
We use the frozen $512$-dimensional SimCLR ResNet-18 features from $5{,}000$ sampled images ($50$ per class) released
with SCAN, standardized per dimension with two levels of $K=(20,100)$.

\textbf{Tasic.}
We use the released PCA-$50$ representation (i.e. $O=I$) of $23{,}822$ mouse cortex
cells from single-cell RNA sequencing with $K=(4,23,133)$.

\textbf{20 Newsgroups.}
We embed the $18{,}846$ posts with the
frozen sentence encoder all-MiniLM-L6-v2 ($384$ dimensions, standardized) with $K=(6,20)$.

\subsection{Baselines}
\label{sec:baselines}

We compare NEST against methods from five clustering families (flat, agglomerative, divisive, hyperbolic, and deep hierarchical clustering). All baselines operate on the same frozen input representation as NEST and are evaluated at the same target cluster counts $K_\ell$.

Flat clustering is represented by $k$-means, run independently at each $K_\ell$. Agglomerative clustering is represented by Ward and average linkage, which each construct a single dendrogram and cut it at the required resolutions. Divisive clustering is represented by bisecting $k$-means, which recursively splits clusters until the target number of clusters is reached. HypHC \citep{HypHC} serves as the hyperbolic hierarchical baseline, and DeepECT \citep{deepect} as the deep hierarchical baseline. We additionally report HDBSCAN \citep{hdbscan} on the synthetic datasets as a density-based baseline. For NEST, each level is evaluated using its learned centers on the corresponding prefix $z_{1:m_\ell}$.

\subsection{Metrics}
\label{sec:eval}

We use three standard clustering metrics at each resolution. ACC is clustering accuracy after one-to-one matching between predicted clusters and ground-truth classes using the Hungarian algorithm \citep{hungarian}. ARI is the adjusted Rand index, which measures pairwise agreement corrected for chance, and NMI is normalized mutual information between predicted and ground-truth partitions.

To evaluate whether the predicted resolutions agree with one another, we also report cross-level consistency. After matching clusters to classes independently at each level, consistency is the fraction of points for which the matched fine-level class has the matched coarse-level class as its parent in the ground-truth labels. We finally compare the training compute and memory requirements of the methods using analytical estimates. Full details are in Appendix~\ref{app:compute}.

\section{Results}
\label{sec:results}

\subsection{Synthetic hierarchical clustering}
\label{sec:synth-results}

We first evaluate NEST as an unsupervised clustering method. Table~\ref{tab:synth} compares NEST with clustering directly in the
observed representation.

Across the synthetic datasets, NEST achieves the strongest overall clustering
performance across resolutions, with the best results at every middle and fine
resolution. Averaged across datasets, NEST improves ACC over the strongest baseline on
each dataset by $0.18$ at the middle resolution and by $0.07$ at the fine
resolution. These gains are also reflected in the permutation-invariant
metrics: averaged across datasets, NEST improves NMI and ARI by $0.13$ and
$0.17$, respectively, at the middle resolution, and by $0.22$ and $0.18$ at
the fine resolution. At the coarse resolution, NEST matches the baselines on Nested GMM, performs best on the balanced tree, and is weaker than Ward on the random-branching tree.

\subsection{Real-data hierarchical clustering}
\label{sec:real-results}

\begin{figure*}[t]
\centering
\includegraphics[width=\textwidth]{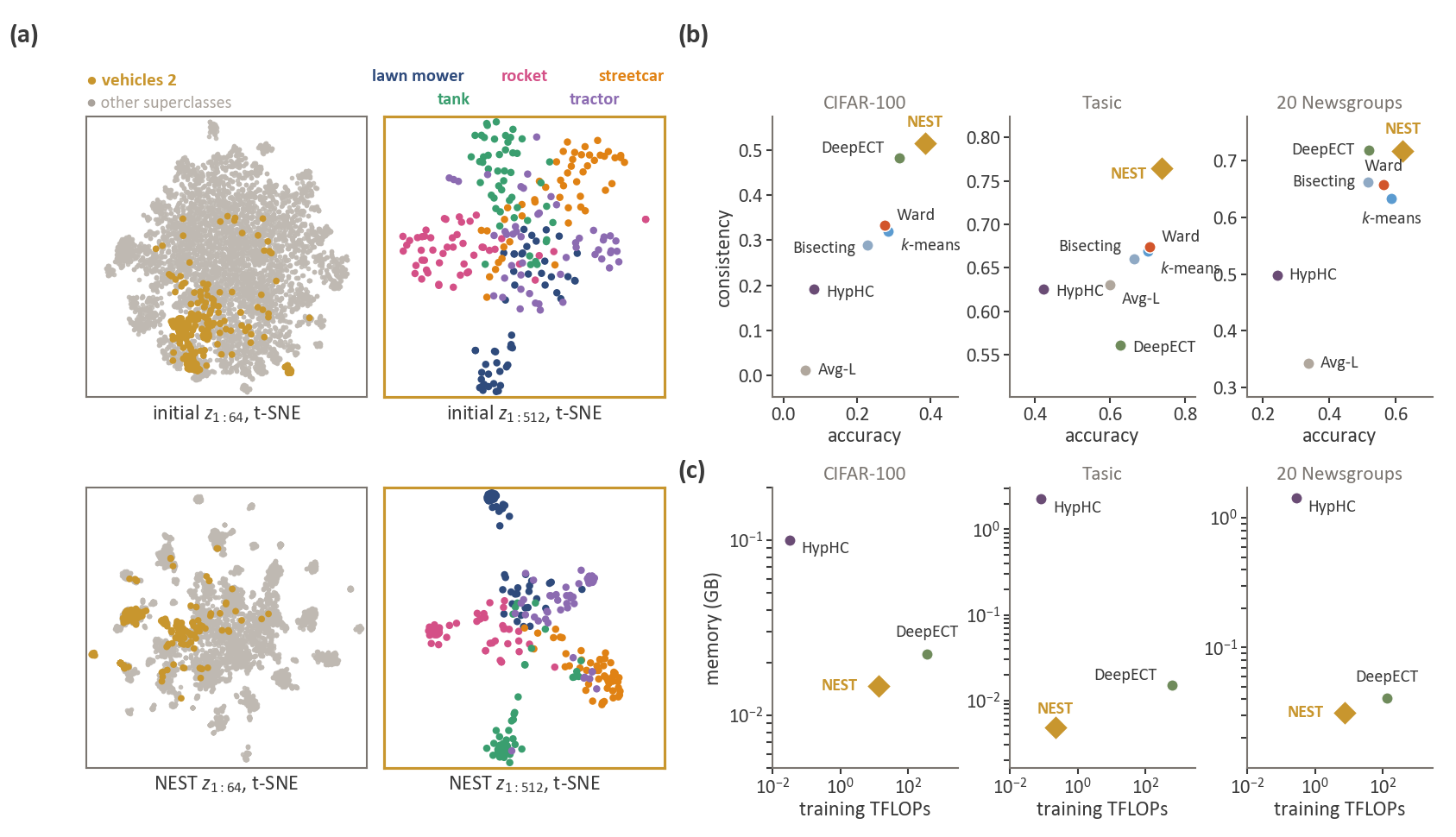}
\vspace{-1cm}
\caption{\textbf{Clustering quality and structure.}
(a) CIFAR-100 vehicles 2. Left: t-SNE of the
coarse prefix $z_{1:64}$ at
initialization (top) and after training (bottom), with the superclass
highlighted in gold. Right: t-SNE of the superclass's image examples in the full embedding $z_{1:512}$, colored by class. All panels: same t-SNE settings and seed.
(b) Mean clustering accuracy versus cross-level consistency.
(c) Training compute versus memory.}
\label{fig:real}
\end{figure*}

We evaluate hierarchical clustering on CIFAR-100, Tasic and 20 Newsgroups
using the same frozen input representation and the same points for all methods.

Figure~\ref{fig:real}a shows the t-SNE~\citep{tsne} visualization before training in the top row and after NEST training in the bottom row. The left panels plot the coarse prefix, showing how NEST reorganizes the embedding at the superclass resolution. The right panels zoom in on the vehicles 2 superclass using the full embedding, where its five constituent classes become substantially more separated after training. This illustrates how NEST keeps the coarse grouping while separating the finer classes; the quantitative evidence is in Table~\ref{tab:real-results} and Figure~\ref{fig:real}b.

We next evaluate clustering quality across resolutions (Table~\ref{tab:real-results}). Overall, NEST achieves strong performance relative to the baselines at most resolutions across all three real datasets. It performs best at both levels on CIFAR-100 (ACC: $.42$ and $.35$) and 20 Newsgroups (ACC: $.68$ and $.57$). On Tasic, it matches centroid-based baselines at the coarse level and performs best at both finer resolutions (ACC: $.78$ at $K=23$, $.66$ at $K=133$). On the coarsest Tasic partition, HypHC reaches $.94$ ACC, while centroid-based methods remain near $.77$, likely because the coarse Tasic partition is highly imbalanced, a setting in which centroid-based objectives may be disadvantaged.

Next, we measure how well each method balances clustering accuracy with cross-level consistency. Figure~\ref{fig:real}b shows that NEST achieves the strongest overall tradeoff across the three real datasets. It shows both the highest mean accuracy and consistency on CIFAR-100 and Tasic, while on 20 Newsgroups it matches the highest consistency with substantially better accuracy. This indicates that NEST improves clustering quality without sacrificing agreement across resolutions.

Finally, Figure~\ref{fig:real}c compares computational cost. NEST requires substantially less training compute and memory than DeepECT and HypHC, reflecting the efficiency of learning a residual transformation on top of frozen features rather than training a full deep learning or hyperbolic hierarchical model.
\subsection{Hierarchy distance in the learned embedding}
\label{sec:tree-distance}

For a pair of points $(i,j)$, let the hierarchy contain $L$ clustering
levels, indexed from coarse ($\ell=1$) to fine ($\ell=L$), and let
$\ell^\ast$ be the first level at which the two points belong to different
clusters, i.e., the level just below their lowest common ancestor. We define
the hierarchy distance
\[
d_T(i,j)=L-\ell^\ast+1,
\]
the number of levels at which the two points are separated. For example, in a
three-level hierarchy, two points that first separate at level $1$ have
distance $3$, whereas points that first separate at level $3$ have distance
$1$; points with the same finest-level label have distance $0$. This is a
discrete distance derived from the tree, monotone in the depth of the lowest
common ancestor, rather than the path length between leaves.

We compare $d_T$ with a multiscale distance in the embedding that has the same
structure, one term per level,
\[
d_z(i,j)=\sum_{\ell=1}^{L}\frac{a_\ell}{\widehat s_\ell}\bigl\|z_i^{(\ell)}-z_j^{(\ell)}\bigr\|_2 ,
\]
where each term uses the level-$\ell$ prefix with the same scaling as
Eq.~\eqref{eq:calibrated-distance} (Appendix~\ref{app:tree}). Because the
prefixes are nested, a difference in the coarse coordinates enters every term,
just as a pair that separates at the coarsest level is counted at every level
by $d_T$. This emphasis on coarse structure is intended, as a coarse split should contribute more to the distance than a split at a finer level. $d_z$ is used only for evaluation.

We evaluated the Pearson correlation between $d_z$ and $d_T$ at initialization
and after training. The correlation increases on every dataset, with particularly clear gains on CIFAR-100 ($.20$ to $.43$) and balanced tree ($.78$ to $.91$), and more modest increases on random branching ($.83$ to $.90$), Tasic ($.86$ to $.90$), 20 Newsgroups ($.42$ to $.46$), and Nested GMM ($.87$ to $.88$). Overall, training makes the learned embedding's multiscale geometry more consistent with the underlying hierarchy.

\subsection{Ablation}
\label{sec:ablation}

To test the effectiveness of each objective in NEST, we removed each component one at a time from the shared configuration and reported the resulting performance in Table~\ref{tab:ablation}. Removing the clustering loss, the separation loss, or the distance scaling consistently reduces ACC and NMI across both CIFAR-100 and Tasic. On CIFAR-100, for example, class-level ACC drops from $.35$ to $.32$ without $\mathcal{L}_{\mathrm{DEC}}$, to $.25$ without $\mathcal{L}_{\mathrm{sep}}$, and to $.30$ without the distance scaling ($a_\ell/\widehat s_\ell=1$, with the margin $\rho_\ell$ kept); similar drops appear at the superclass level and on Tasic.

The two anti-collapse regularizers matter when their corresponding failure modes occur. Removing $\mathcal{L}_{\mathrm{rank}}$ hurts CIFAR-100, reducing ACC from $.42$ to $.28$ at $K=20$ and from $.35$ to $.11$ at $K=100$. This degradation coincides with a drop in effective rank from $131$--$136$ to $5$--$8$. Tasic does not exhibit this rank collapse; instead, its main failure mode is excessive contraction at the finest resolution.
We applied $\mathcal{L}_{\mathrm{within}}$ only at the finest resolution on the real datasets (Table~\ref{tab:hyper}).

\begin{table}[t]
\centering
\small
\setlength{\tabcolsep}{5pt}
\caption{\textbf{Ablation.} Each row removes one component. Metrics mean$\pm$s.d.\ over $3$ seeds.}
\label{tab:ablation}
\begin{tabular}{l cc}
\toprule
Variant & ACC & NMI \\
\midrule

\multicolumn{3}{l}{\textit{CIFAR-100, $K{=}20$}} \\
NEST
& $\mathbf{.42}_{\pm.01}$ & $\mathbf{.43}_{\pm.01}$ \\
$-\mathcal{L}_{\mathrm{sep}}$
& $.38_{\pm.01}$ & $.39_{\pm.00}$ \\
$-\mathcal{L}_{\mathrm{DEC}}$
& $.28_{\pm.01}$ & $.30_{\pm.01}$ \\
$-\mathcal{L}_{\mathrm{rank}}$
& $.28_{\pm.05}$ & $.36_{\pm.05}$ \\
$-\mathcal{L}_{\mathrm{within}}$
& $.41_{\pm.01}$ & $\mathbf{.43}_{\pm.01}$ \\
No distance scaling
& $.36_{\pm.01}$ & $.38_{\pm.01}$ \\

\midrule
\multicolumn{3}{l}{\textit{CIFAR-100, $K{=}100$}} \\
NEST
& $\mathbf{.35}_{\pm.01}$ & $\mathbf{.56}_{\pm.00}$ \\
$-\mathcal{L}_{\mathrm{sep}}$
& $.25_{\pm.00}$ & $.48_{\pm.00}$ \\
$-\mathcal{L}_{\mathrm{DEC}}$
& $.32_{\pm.00}$ & $.52_{\pm.00}$ \\
$-\mathcal{L}_{\mathrm{rank}}$
& $.11_{\pm.01}$ & $.36_{\pm.04}$ \\
$-\mathcal{L}_{\mathrm{within}}$
& $\mathbf{.35}_{\pm.01}$ & $\mathbf{.56}_{\pm.00}$ \\
No distance scaling
& $.30_{\pm.01}$ & $.50_{\pm.00}$ \\

\midrule
\multicolumn{3}{l}{\textit{Tasic, $K{=}133$}} \\
NEST
& $\mathbf{.66}_{\pm.01}$ & $\mathbf{.85}_{\pm.00}$ \\
$-\mathcal{L}_{\mathrm{sep}}$
& $.54_{\pm.01}$ & $.80_{\pm.00}$ \\
$-\mathcal{L}_{\mathrm{DEC}}$
& $.56_{\pm.01}$ & $.84_{\pm.00}$ \\
$-\mathcal{L}_{\mathrm{rank}}$
& $\mathbf{.66}_{\pm.01}$ & $\mathbf{.85}_{\pm.00}$ \\
$-\mathcal{L}_{\mathrm{within}}$
& $.57_{\pm.01}$ & $.83_{\pm.00}$ \\
No distance scaling
& $.56_{\pm.02}$ & $.81_{\pm.01}$ \\

\bottomrule
\end{tabular}
\vspace{-12pt}
\end{table}

\section{Conclusion}
\label{sec:conclusion}
We introduced NEST, a representation-learning method that encodes several
clustering resolutions in nested prefixes of a single embedding and trains
them jointly with per-level clustering, center-separation, and anti-collapse
objectives. On synthetic hierarchies, NEST demonstrated large clustering gains at middle and finer resolutions. Distances in the learned embedding are also more strongly correlated with the hierarchy distance than distances in the input embedding. On real datasets spanning image, text, and single-cell modalities, NEST shows consistent gains over strong baselines at most clustering resolutions, particularly on CIFAR-100, 20 Newsgroups, and the finer Tasic levels.

Several challenging directions remain. Highly imbalanced hierarchies remain
difficult for centroid-based objectives. More broadly, open problems in
hierarchical clustering include automatically determining the number of
clusters and resolutions, handling irregular or partially observed
hierarchies, and scaling nested clustering to larger multimodal
representations.

\subsubsection*{Acknowledgments}
All authors declare no competing interests.

This work was supported by the National Science Foundation under Grant CIF-2403452.

\subsubsection*{AI Use Statement}
Generative AI tools were used for language editing, code assistance, and figure refinement. All methods, experiments, analyses, and conclusions were developed and verified by the authors.

\clearpage
\bibliographystyle{plainnat}
\bibliography{ref}

\newpage
\appendix

\section{Datasets}
\label{app:data}

\paragraph{Synthetic generators.}
Each of the three synthetic datasets has a three-level hierarchy. Each
level is encoded in a separate coordinate block before a fixed random
orthonormal transformation is applied, so the hierarchy is not axis-aligned
in the observed space. Two parameters control difficulty. The separation
factor $\phi$ sets the distance between sibling cluster centers relative to
the within-cluster spread; $\phi\approx5$ keeps siblings well separated,
while $\phi\approx2.5$ makes them overlap. The scale-separation factor $\kappa\in(0,1)$
controls the relative magnitude of successive hierarchical levels,
$\mathrm{disp}_\ell=\mathrm{disp}_L\kappa^{-(L-\ell)}$; smaller $\kappa$
produces well-separated scales, whereas $\kappa\to1$ makes distinctions
across levels comparable in magnitude. The generators below are common to
every synthetic run; the real-data model settings are in Table~\ref{tab:hyper}.

Nested GMM: $K=(4,16,64)$, $\mathrm{obs\_dim}=99$, data blocks $(33,33,33)$, branching $4$,
$\texttt{base\_scale}=8.0$, $\kappa=0.4$, $r_{\mathrm{target}}=0.55$, $200$ points/leaf.
Balanced tree: $K=(3,7,15)$, $\mathrm{obs\_dim}=20$, blocks $(7,7,6)$, $\phi=4$,
$\kappa=0.6$, $80$ points/node. Random-branching tree: $K=(4,13,52)$, $1+\mathrm{Poisson}(3)$ children per node (at most $6$),
depth $4$, $\phi=4$, $\kappa=0.6$, $50$ points/node, topology fixed across seeds. In all
three the clean data is mapped through a fixed random orthonormal rotation and augmented
with $50$ nuisance coordinates of standard deviation $\sigma_{\mathrm{nu}}=1$, giving
$149$ observed dimensions for the GMM and $70$ for the two trees. Geometry and sampling noise are redrawn per seed at fixed
cluster counts.

\paragraph{CIFAR-100.}
We use the frozen $512$-dimensional SimCLR ResNet-18 pretext features
released with SCAN \citep{scan}: the CIFAR-100 pretext checkpoint
\texttt{simclr\_cifar100.pth.tar} from the authors' repository
(\texttt{wvangansbeke/Unsupervised-Classification}), read at the backbone
output before the $128$-dimensional projection head, with SCAN's image
normalization. Features of all $60{,}000$ train and test images are
standardized per dimension with statistics computed over those $60{,}000$;
no labels or self-labeling are used. The $20$ superclasses and $100$ classes define
$K=(20,100)$. Every method, NEST included, is fit on the same fixed subset
of $5{,}000$ images drawn as $50$ per class (seed $0$), and scored on
all of them (evaluation points, below).

\paragraph{Tasic et al.\ 2018.}
We use the released preprocessed scRNA-seq representation from \cite{tasic2018shared}
(library-size normalization, $\log_2$ transformation,
highly-variable-gene selection, and PCA to $50$ dimensions) for
$23{,}822$ cells, the matrix \texttt{tasic-pca50.npy} distributed with
the berenslab single-cell visualization code, used raw (not
standardized). Labels come from the Allen Institute 2018-06-14 release
of the mouse VISp and ALM data
(\texttt{mouse\_VISp\_2018-06-14\_samples-columns.csv},
\texttt{mouse\_ALM\_2018-06-14\_samples-columns.csv}). The official Allen \texttt{class}, \texttt{subclass},
and \texttt{cluster} annotations define $K=(4,23,133)$. Samples are
aligned to the released PCA representation using \texttt{sample\_name};
the join is one-to-one with no missing cells. The resulting taxonomy is
strictly nested: each cluster belongs to one subclass and each subclass
to one class. The official four-way class annotation is used throughout.
For sensitivity analysis only, we also consider the broader three-way
grouping obtained by merging Endothelial with Non-Neuronal cells.

\paragraph{20 Newsgroups.} 20 Newsgroups \citep{20news} is a text classification dataset containing approximately 20,000 documents organized into 20 topic categories. The $18{,}846$ posts of the scikit-learn \texttt{20newsgroups} ``all'' set,
with headers, footers and quotes removed, are embedded with the frozen
sentence encoder \texttt{sentence-transformers/all-MiniLM-L6-v2} \citep{sentencebert, minilm} (maximum
sequence length $256$, $384$ dimensions) and standardized per dimension.
The $20$ newsgroups define the fine level; grouping them by name gives the
$6$ coarse topics comp ($5$ groups), rec ($4$), sci ($4$), politics ($3$),
religion ($3$: alt.atheism, soc.religion.christian, talk.religion.misc) and
forsale ($1$), $K=(6,20)$. Coarse classes hold $975$ to $4{,}891$ posts.
\paragraph{Evaluation.}
On the datasets, every method is fit on all points without labels and
scored on all points ($5{,}000$ images for CIFAR-100, $23{,}822$ cells for
Tasic, $18{,}846$ posts for 20 Newsgroups). Labels are not used for training, initialization, or stopping; they are used only to construct the class-balanced CIFAR-100 subset and for evaluation.

\section{Geometric motivation for the \texorpdfstring{$4r_\ell$}{4r} separation margin}
\label{app:margin-geometry}

In Section~\ref{sec:scale-calibration}, we set the separation margin to
$\rho_\ell = 4\widehat r_\ell$. The factor of $4$ is motivated by the following geometric argument.

Let $r_\ell$ denote an upper bound on the distance from any point in a cluster
to its center at level $\ell$, and let $s_\ell$
denote the distance between two cluster centers. Since each point can lie up
to $r_\ell$ from its cluster center, two points within the same cluster can be
as far as $2r_\ell$ apart.

For two different clusters, the closest pair of points occurs when both points
lie toward the other cluster, giving a minimum cross-cluster distance of
\[
s_\ell - 2r_\ell.
\]
Requiring this distance to be at least the maximum within-cluster distance gives
\[
s_\ell - 2r_\ell \ge 2r_\ell,
\]
making
\[
s_\ell \ge 4r_\ell.
\]
In real data, the maximum point-to-center distance is often determined by a
small number of outlying points. We therefore use the mean point-to-center
distance at initialization, $\widehat r_\ell$, as a robust estimate of the
cluster scale. Motivated by the geometric bound above, we set
\[
\rho_\ell = 4\widehat r_\ell.
\]
Because $\widehat r_\ell$ is a mean rather than an upper bound, the inequality
above is no longer guaranteed for every point; the factor $4$ is a heuristic
taken from the worst case.

\section{Baseline settings}
\label{app:baselines}
$k$-means uses \texttt{n\_init=20}.
Ward and average linkage construct a single dendrogram that is cut at
each target $K$. Average linkage uses Euclidean distance: every method in
Table~\ref{tab:real-results}, NEST included, then operates on the same
Euclidean geometry of the frozen input, so differences reflect the
clustering procedure rather than a change of metric on the input.
HDBSCAN uses
$\texttt{min\_cluster\_size}=\max(5,N/2k)$, $\texttt{min\_samples}=5$,
EOM selection, with noise points counted as one label.
HypHC uses rank-$2$ embeddings, $50$ epochs of Riemannian Adam, and
greedy tree decoding, with $504{,}000$ triplets for CIFAR-100 and
20 Newsgroups and $2{,}000{,}000$ for Tasic.
DeepECT uses its published configuration: an autoencoder
($D$--$500$--$500$--$2000$--$10$) pretrained for $50{,}000$ steps, after which
the cluster tree is grown by splitting the leaf with the largest
within-cluster error every $500$ steps until it has as many leaves as the
finest level, followed by $20{,}000$ further steps; the flat partition at
each $K$ is the tree as it stood when it had $K$ leaves.
On every dataset each NEST level is read out by its own centers on the
prefix $z_{1:m_\ell}$, while the baselines run on the input.
Stochastic methods are reported over seeds $\{0,1,2\}$ on real data and
over $5$ seeds on synthetic data.

\section{Metric definitions}
\label{app:metrics}
ACC aligns clusters to classes by the Hungarian algorithm \citep{hungarian}. With $K$
predicted clusters and $K$ true classes at a level, we form the
$K\times K$ matrix of class--cluster co-occurrence counts and take the
one-to-one assignment that maximizes the total count on the diagonal;
ACC is that total divided by $n$. Wherever a cluster is called the one
\emph{matched} to a class, this is the assignment meant. Because it is
one-to-one, a class whose points fall mostly into a cluster already
claimed by another class is matched to whatever cluster remains, possibly
an empty one; ARI and NMI do not use this matching.

\paragraph{Cross-level consistency.}
ACC, ARI and NMI score each level separately, so a method can score well at
every level while its levels disagree about the hierarchy, for example by
placing a point in a coarse cluster matched to \emph{vehicles~1} but in a fine
cluster matched to \emph{apple}. Cross-level consistency measures this
agreement. For each pair of adjacent levels, we match each level's clusters to
its true classes with the Hungarian assignment used for ACC. Each point $i$
then has a predicted coarse class $\hat c_i$, the class matched to its coarse
cluster, and a predicted fine class $\hat f_i$, the class matched to its fine
cluster. Point $i$ is \emph{consistent} when the true parent of $\hat f_i$ is
$\hat c_i$, that is, $\mathrm{par}(\hat f_i)=\hat c_i$. The consistency of a
level pair is the fraction of consistent points, and a dataset's consistency is
the mean over its adjacent level pairs. Consistency does not require
correctness: a point is consistent even if both of its predicted classes are
wrong, as long as they agree with the true hierarchy. A method that cuts one
dendrogram at both levels produces nested clusters by construction, but it need
not be fully consistent, because each level is matched to its classes
separately.

\section{Hierarchy-distance correlation}
\label{app:tree}
We measure how well pairwise distances in the learned embedding reflect the
ground-truth hierarchy. Equation~\eqref{eq:calibrated-distance} defines the
scaled distance between a data point and a cluster center at level $\ell$.
Here, we use the same scaling to measure the distance between two data points.
For a pair $(i,j)$, we define
\[
d_z(i,j)
=
\sum_{\ell=1}^{L}
\frac{a_\ell}{\widehat{s}_\ell}
\left\|
z_i^{(\ell)}-z_j^{(\ell)}
\right\|_2,
\]
where $z_i^{(\ell)}$ and $z_j^{(\ell)}$ are the level-$\ell$ prefixes of the
two points.

Let $\ell^\ast$ denote the coarsest level at which the ground-truth cluster
labels of $i$ and $j$ differ. We define
\[
d_T(i,j)=L-\ell^\ast+1,
\]
the number of levels at which the two points are separated, so pairs that separate earlier in the hierarchy have larger hierarchy distance. Because the prefixes are nested, a difference in the coarse coordinates enters every term of $d_z$, in the same way that a coarse split is counted at every level by $d_T$; $d_z$ is used only for evaluation.

We sample $2{,}000$ pairs for each distinct value of $d_T$ and compute the
Pearson correlation between $d_z$ and $d_T$ over the resulting pairs.

\section{Training compute and memory}
\label{app:compute}

Figure~\ref{fig:real}c reports training compute in floating-point operations
(FLOPs) and memory. FLOPs count the multiplications and additions of a
method's full training run: one forward pass through each layer, tripled to
include the backward pass, summed over every training sample and step.
Initialization, optimizer updates and evaluation are not counted. Memory is the
space needed to hold the training data and the learned parameters in 32-bit
floats.

HypHC's FLOPs and memory are computed differently from those of NEST and
DeepECT, which are counted over the individual training points they process.
HypHC instead trains on triplets of points: it learns a hierarchy by minimizing
a continuous version of Dasgupta's cost, which is a sum over triplets, where
for each triplet the two most similar points should be merged lowest in the
tree. Using every triplet is infeasible, since their number grows with the cube
of the number of points (about $2\times10^{10}$ for the $5{,}000$ CIFAR-100
images), so HypHC is trained on a fixed random sample of triplets: $504{,}000$
for CIFAR-100 and 20 Newsgroups and $2{,}000{,}000$ for Tasic, for $50$ epochs
each. Its FLOPs are counted over these triplets and also include building the
full matrix of pairwise similarities and decoding the tree, both of which grow
with the square of the number of points. Its memory is dominated by that
similarity matrix, rather than by the training data and parameters as for NEST
and DeepECT.

\section{Single-level controls}
\label{app:single}

To separate the effect of training the levels jointly from the effect of the
NEST objective itself, we trained each level on its own in two ways: NEST with a
single head on the full embedding, every other setting of the shared
configuration unchanged, and single-scale DEC (autoencoder
$D\!\to\!512\!\to\!256\!\to\!64$, $100$ warm-up epochs, $160$ DEC epochs) at each
$K_\ell$. Both were fit and scored on all points over seeds $0$, $1$ and $2$ and
compared with the jointly trained Table~\ref{tab:real-results} model. We found
that joint training raises the coarsest level on every dataset and is above or
within one standard deviation of single-level NEST elsewhere, and that
single-scale DEC is below NEST except at the two coarser Tasic levels, where it
is within one standard deviation (Table~\ref{tab:single}).

\begin{table*}[t]
\centering
\scriptsize
\caption{\textbf{Joint versus single-level training.} ACC\,/\,ARI\,/\,NMI per level, mean$\pm$s.d.\ over $3$ seeds, all points scored. Bold is the best of the three after rounding, ties share it.}
\label{tab:single}
\setlength{\tabcolsep}{3pt}
\resizebox{\textwidth}{!}{%
\begin{tabular}{@{}ll ccc ccc ccc@{}}
\toprule
& & \multicolumn{3}{c}{NEST, joint} & \multicolumn{3}{c}{NEST, one level} & \multicolumn{3}{c}{DEC, one level} \\
\cmidrule(lr){3-5}\cmidrule(lr){6-8}\cmidrule(lr){9-11}
Dataset & $K$ & ACC & ARI & NMI & ACC & ARI & NMI & ACC & ARI & NMI \\
\midrule
\multirow{2}{*}{CIFAR-100}
& $20$
& \textbf{.42}{\scriptsize\textpm.01}
& \textbf{.24}{\scriptsize\textpm.01}
& \textbf{.43}{\scriptsize\textpm.01}
& .38{\scriptsize\textpm.02}
& .20{\scriptsize\textpm.02}
& .38{\scriptsize\textpm.01}
& .34{\scriptsize\textpm.01}
& .17{\scriptsize\textpm.00}
& .39{\scriptsize\textpm.00} \\

& $100$
& .35{\scriptsize\textpm.01}
& \textbf{.22}{\scriptsize\textpm.00}
& \textbf{.56}{\scriptsize\textpm.00}
& \textbf{.36}{\scriptsize\textpm.01}
& \textbf{.22}{\scriptsize\textpm.01}
& .55{\scriptsize\textpm.01}
& .27{\scriptsize\textpm.01}
& .11{\scriptsize\textpm.01}
& .51{\scriptsize\textpm.00} \\

\midrule
\multirow{2}{*}{20 Newsgroups}
& $6$
& \textbf{.68}{\scriptsize\textpm.01}
& \textbf{.44}{\scriptsize\textpm.02}
& \textbf{.47}{\scriptsize\textpm.02}
& .51{\scriptsize\textpm.00}
& .30{\scriptsize\textpm.00}
& .40{\scriptsize\textpm.00}
& .57{\scriptsize\textpm.05}
& .38{\scriptsize\textpm.01}
& .40{\scriptsize\textpm.01} \\

& $20$
& \textbf{.57}{\scriptsize\textpm.02}
& \textbf{.41}{\scriptsize\textpm.01}
& \textbf{.54}{\scriptsize\textpm.01}
& .56{\scriptsize\textpm.02}
& .40{\scriptsize\textpm.02}
& \textbf{.54}{\scriptsize\textpm.01}
& .52{\scriptsize\textpm.01}
& .35{\scriptsize\textpm.00}
& .51{\scriptsize\textpm.01} \\

\midrule
\multirow{3}{*}{Tasic 2018}
& $4$
& \textbf{.77}{\scriptsize\textpm.00}
& \textbf{.79}{\scriptsize\textpm.00}
& \textbf{.83}{\scriptsize\textpm.00}
& .72{\scriptsize\textpm.01}
& .62{\scriptsize\textpm.01}
& .61{\scriptsize\textpm.01}
& .76{\scriptsize\textpm.02}
& .76{\scriptsize\textpm.06}
& .79{\scriptsize\textpm.06} \\

& $23$
& \textbf{.78}{\scriptsize\textpm.01}
& .74{\scriptsize\textpm.01}
& \textbf{.89}{\scriptsize\textpm.00}
& \textbf{.78}{\scriptsize\textpm.01}
& .74{\scriptsize\textpm.00}
& \textbf{.89}{\scriptsize\textpm.00}
& \textbf{.78}{\scriptsize\textpm.02}
& \textbf{.75}{\scriptsize\textpm.02}
& \textbf{.89}{\scriptsize\textpm.00} \\

& $133$
& \textbf{.66}{\scriptsize\textpm.01}
& \textbf{.64}{\scriptsize\textpm.01}
& \textbf{.85}{\scriptsize\textpm.00}
& .63{\scriptsize\textpm.01}
& .61{\scriptsize\textpm.01}
& \textbf{.85}{\scriptsize\textpm.00}
& .56{\scriptsize\textpm.01}
& .50{\scriptsize\textpm.00}
& .84{\scriptsize\textpm.00} \\

\bottomrule
\end{tabular}}
\end{table*}

\section{Hyperparameters}
\label{app:hyper}

Table~\ref{tab:hyper} shows the hyperparameters we used for each dataset.

\begin{table*}[t]
\centering
\small
\caption{\textbf{Hyperparameters of NEST}}
\label{tab:hyper}
\footnotesize
\setlength{\tabcolsep}{3pt}
\resizebox{\textwidth}{!}{%
\begin{tabular}{@{}l ccc ccc@{}}
\toprule
& CIFAR-100 & Tasic 2018 & 20 Newsgroups & Balanced tree & Nested GMM & Random-branching \\
\midrule
\multicolumn{7}{@{}l}{\emph{Input}} \\
Representation & SCAN SimCLR & PCA & MiniLM & raw & raw & raw \\
Preprocessing & standardized & raw & standardized & -- & -- & -- \\
$D$ & $512$ & $50$ & $384$ & $70$ & $149$ & $70$ \\
$K_\ell$ & $20/100$ & $4/23/133$ & $6/20$ & $3/7/15$ & $4/16/64$ & $4/13/52$ \\
\midrule
\multicolumn{7}{@{}l}{\emph{Model}} \\
Coordinates (basis $O$) & principal axes & principal axes (released PCA) & principal axes & native & native & native \\
Prefix widths $m_\ell$ & $64/512$ & $6/25/50$ & $32/384$ & $8/35/70$ & $18/74/149$ & $8/35/70$ \\
Residual scale $\gamma$ & $0.1$ & $0.1$ & $0.1$ & $0.3$ & $0.3$ & $0.3$ \\
\midrule
\multicolumn{7}{@{}l}{\emph{Objective}} \\
Level weights $a_\ell$ & $0.5{:}1$ & $0.5{:}1{:}1$ & $0.5{:}1$ & $1{:}1{:}1$ & $1{:}2{:}4$ & $1{:}1{:}1$ \\
$\lambda_{\mathrm{rank}}$ & $10$ & $10$ & $0.3$ & $0.1$ & $4.3$ & $0.1$ \\
$\tau_{\mathrm r}$ (rank fraction) & $0.75$ & $0.75$ & $0.75$ & $0.75$ & $0.75$ & $0.75$ \\
$\lambda_{\mathrm{within}}$ & $10$ & $10$ & $10$ & $0.1$ & $4.3$ & $0.1$ \\
$\tau_{\mathrm w}$ (within fraction) & $0.75$ & $0.75$ & $0.75$ & $0.75$ & $0.75$ & $0.75$ \\
Levels where $\mathcal{L}_{\mathrm{within}}$ acts & finest & finest & finest & all & finest & all \\
\bottomrule
\end{tabular}}
\end{table*}

\section{Sensitivity analysis}
\label{app:sensitivity}

\begin{figure*}[t]
\centering
\includegraphics[width=\textwidth]{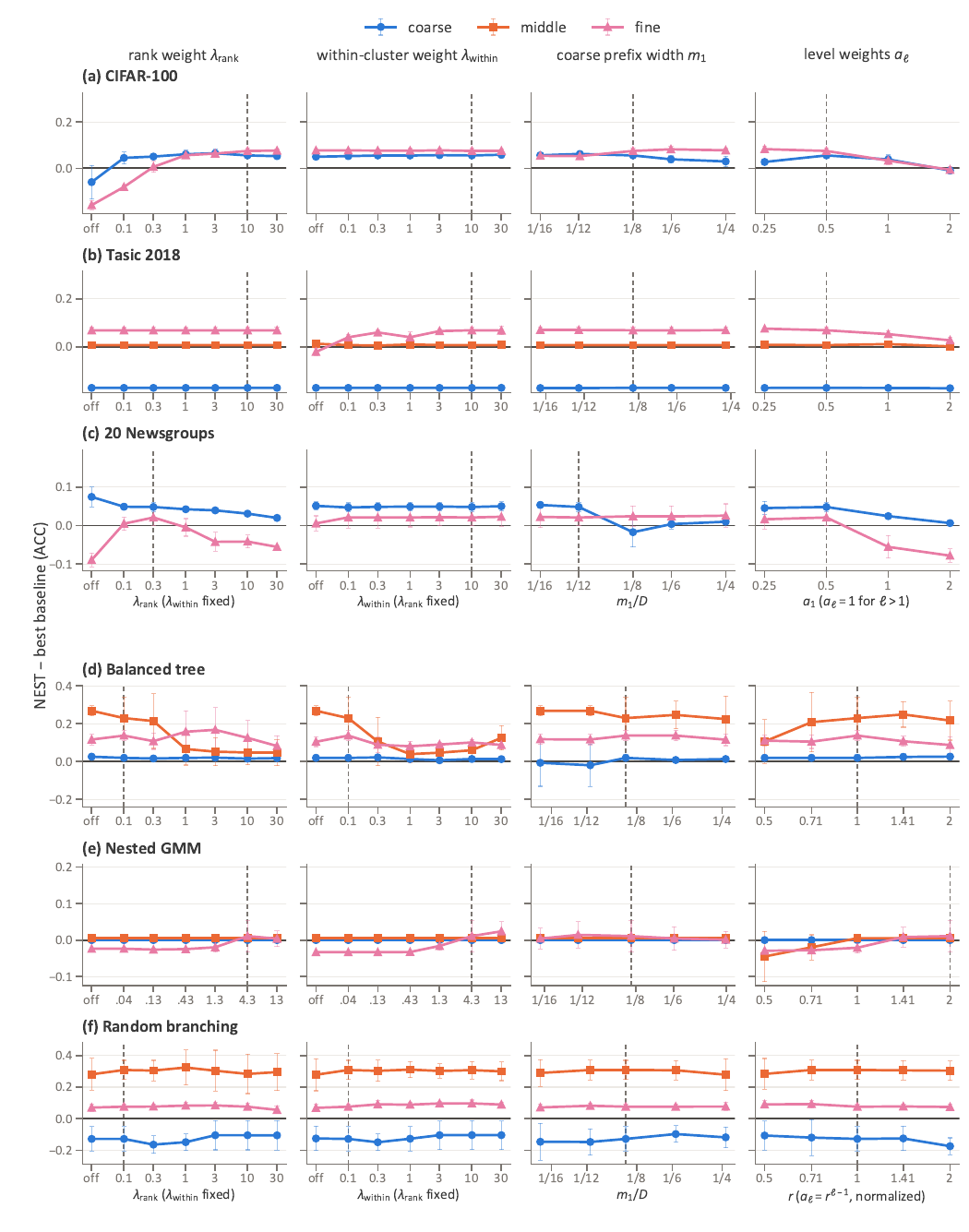}
\caption{\textbf{Sensitivity analysis of NEST.} Accuracy of NEST minus that
of the best baseline per level, varying one setting at a time; mean $\pm$
s.d.\ over $3$ (real) or $5$ (synthetic) seeds. Columns: $\lambda_{\mathrm{rank}}$
with $\lambda_{\mathrm{within}}$ fixed; $\lambda_{\mathrm{within}}$ with
$\lambda_{\mathrm{rank}}$ fixed; coarse prefix width $m_1$; level weights (real
data: coarse weight $a_1$, others $1$; synthetic: $a_\ell=r^{\ell-1}$ with the
regularization weights fixed). Dashed line: reported configuration.}
\label{fig:sensitivity}
\end{figure*}

As in other unsupervised clustering settings, hyperparameter selection is
challenging because ground-truth labels are unavailable for validation. We
therefore limit dataset-specific tuning and share most settings across datasets
(Table~\ref{tab:hyper}). The few dataset-specific settings are selected from a
small number of variants evaluated on the reported data.
Figure~\ref{fig:sensitivity} varies each of these settings individually around
the reported value and shows similar performance across a range of nearby values.

Figure~\ref{fig:sensitivity} changes one setting at a time around the reported
configuration: the anti-collapse regularization weights
$\lambda_{\mathrm{rank}}$ and $\lambda_{\mathrm{within}}$ (each varied with the
other fixed), the coarse prefix
width $m_1$, and the relative level weights $a_\ell$. Each point reports the
difference in ACC between NEST and the strongest baseline at the corresponding
level, evaluated on all points.

On CIFAR-100, NEST remains above the strongest baseline at both resolutions
for $\lambda_{\mathrm{rank}}\ge 1$, for every tested $\lambda_{\mathrm{within}}$,
for coarse-prefix widths between $D/16$ and $D/4$, and for $a_1\le 1$. Removing
$\mathcal{L}_{\mathrm{rank}}$ primarily degrades the fine, class-level clustering.
Tasic is comparatively insensitive to $\lambda_{\mathrm{rank}}$ and $m_1$; removing
$\mathcal{L}_{\mathrm{within}}$ mainly hurts the finest level. On 20 Newsgroups, performance remains similar across a range of nearby settings, with the finest level remaining above baseline for $\lambda_{\mathrm{rank}}\leq 0.3$ and for every tested $\lambda_{\mathrm{within}}$, while the coarse-level margin varies with prefix width and becomes smaller near $D/8$.

The synthetic datasets show a similar pattern. On the balanced and
random-branching trees, the middle and finest levels remain above the
baselines throughout the tested ranges. The balanced-tree coarse level also
remains above baseline across most tested prefix widths.
Nested GMM shows greater variation at the finest level, where improvements appear once $\lambda_{\mathrm{rank}}$ or $\lambda_{\mathrm{within}}$ reaches the reported value and with the $1{:}2{:}4$ level weighting.

\end{document}